\documentclass{bmvc2k}

\title{Mining Discriminative Food Regions for Accurate Food Recognition}

\addauthor{Jianing Qiu}{jianing.qiu17@imperial.ac.uk}{1,2}
\addauthor{Frank P.-W. Lo}{po.lo15@imperial.ac.uk}{1}
\addauthor{Yingnan Sun}{y.sun16@imperial.ac.uk}{1,2}
\addauthor{Siyao Wang}{s.wang18@imperial.ac.uk}{1}
\addauthor{Benny Lo}{benny.lo@imperial.ac.uk}{1}
\addinstitution{
 The Hamlyn Centre\\
 Imperial College London\\
 London, UK
}
\addinstitution{
 Department of Computing\\
 Imperial College London\\
 London, UK
}

\runninghead{Qiu et al.}{Mining Discriminative Food Regions}

\def\eg{\emph{e.g}\bmvaOneDot}

\def\etal{\emph{et al}\bmvaOneDot}

\def\ie{\emph{i.e}\bmvaOneDot}

\usepackage[ruled, vlined, linesnumbered]{algorithm2e}
\usepackage{booktabs}
\usepackage{multirow}
\usepackage{graphicx}
\usepackage{url}
\begin{document}

\maketitle

\begin{abstract}
Automatic food recognition is the very first step towards passive dietary monitoring. In this paper, we address the problem of food recognition by mining discriminative food regions. Taking inspiration from \textit{Adversarial Erasing}, a strategy that progressively discovers discriminative object regions for weakly supervised semantic segmentation, we propose a novel network architecture in which a primary network maintains the base accuracy of classifying an input image, an auxiliary network adversarially mines discriminative food regions, and a region network classifies the resulting mined regions. The global (the original input image) and the local (the mined regions) representations are then integrated for the final prediction. The proposed architecture denoted as PAR-Net is end-to-end trainable, and highlights discriminative regions in an online fashion. In addition, we introduce a new fine-grained food dataset named as Sushi-50, which consists of 50 different sushi categories. Extensive experiments have been conducted to evaluate the proposed approach. On three food datasets chosen (Food-101, Vireo-172, and Sushi-50), our approach performs consistently and achieves state-of-the-art results (top-1 testing accuracy of $90.4\%$, $90.2\%$, $92.0\%$, respectively) compared with other existing approaches. Dataset and code are available at~\url{https://github.com/Jianing-Qiu/PARNet}


\end{abstract}

\section{Introduction}
\label{sec:intro}

Diet-induced diseases are becoming increasingly prevalent among populations. One underlying factor is people's poor management of their daily dietary intake. The other factor is that there is currently no accurate measurement of dietary intake. Dietary measurement in nutritional epidemiology is heavily based on self-reported data that are highly inaccurate and subjective~\cite{shim2014dietary}, which hinders nutritionists from designing effective dietary guidance. To mitigate the problem of existing dietary measurement techniques that require extensive user input and produce unsatisfactory results, the concept of passive dietary monitoring is proposed~\cite{passivedietaryintake2017}, which relies on sensors such as cameras to pervasively record eating episodes and automatically perform food recognition, volume estimation, and deduce dietary intake. In realising this concept of passive monitoring, food recognition is the first and a crucial step as any misrecognition will lead to inaccurate measurements afterwards. With recent advances in computer vision, recognising pictured dishes have achieved promising results but still remains as a challenging field of research given that there are enormous varieties of dishes and even the same type of food can have very different appearances. In this work, we aim to achieve accurate food recognition by mining discriminative regions of a food image. This is motivated by the previous work done by Bossard \etal~\cite{bossard2014food} that utilises random forests to mine discriminative components from food images. Unlike~\cite{bossard2014food}, we develop a convolutional neural network (CNN) model and utilise a weakly supervised method to discover discriminative food regions. This weakly supervised method used in both network training and inference is based on \textit{Adversarial Erasing} (AE)~\cite{wei2017object}, a strategy developed for weakly supervised semantic segmentation. One prominent feature of AE is that it enables the discriminative region of an object of interest to be discovered progressively, which in our case enables better recognition of food items. Our implementation of AE however differs from~\cite{wei2017object} in that we integrate it into a new network architecture for object recognition (food recognition in particular), and all sub-networks involved are trained jointly, which is more convenient, compared to its original usage in which networks need to be trained independently. Although region mining is performed, the proposed approach still predicts the food class of an input image efficiently in an end-to-end manner, which will be detailed in Sections~\ref{subsec: cam} and~\ref{subsec: par_net}.

The contributions of our work are twofold: (\romannumeral 1) we propose a new network architecture that is able to mine discriminative food regions in a weakly supervised fashion and be trained end-to-end. The mining strategy is adopted and optimised for food recognition. Comprehensive experiments are performed to validate the proposed approach; (\romannumeral 2) we introduce a new fine-grained food dataset which consists of 50 sub-categories of one common food class, \ie, sushi, in contrast to most existing datasets that only contain coarse food classes.



\section{Related Work}

{\bf Food Recognition.} Existing vision-based approaches for food recognition either use hand-crafted features in conjunction with SVMs, or use CNNs alone. Works in the former such as \cite{bettadapura2015leveraging, chen2009pfid} resort to color and SIFT based features for food image classification. Pairwise local features are developed in~\cite{yang2010food} to capture spatial relationships between the ingredients. In \cite{bossard2014food}, food recognition is decomposed into two steps: first scoring an image's superpixels using the component models trained by random forest mined food parts, and then predicting the food class using a multi-class SVM. The results of these approaches are not satisfactory as neither the learned representations nor the models are capable of distinguishing food items with high intra-class variations. Among the latter CNN-based approaches, Chen and Ngo~\cite{chen2016deep} designed a series of CNN architectures and utilised multi-task learning to simultaneously recognise the food category and ingredients composition. Based on the observation that certain food dishes have a clear vertical layer structure, Martinel \etal~\cite{martinel2018wide} later proposed a slice convolutional layer to learn such information. Integrated with the wide residual network~\cite{Zagoruyko2016WRN}, the resulting architecture shows promising results on food recognition. Some previous works~\cite{beijbom2015menu, bettadapura2015leveraging, meyers2015im2calories} also tried to narrow down the number of food categories by using the restaurant context. Food recognition recently has also been proposed as a topic of challenges in the fine-grained visual categorization competition~\cite{cui2018large, yu2018deep}.

\noindent
{\bf Adversarial Erasing.} 

Adversarial Erasing (AE) is initially introduced by Wei \etal~\cite{wei2017object} aiming to address the weakly supervised semantic segmentation problem. It is an iterative process that enforces a succeeding classification network to discover a new discriminative region from an image with the more discriminative regions removed by the previous networks. Each discriminative region is obtained by thresholding the associated class activation map (CAM)~\cite{zhou2016learning}, which is a visualisation map that highlights the areas in a given image that a classification network relies on for identifying the target class. By merging these mined regions, the object of interest can then be well segmented. To accomplish this, a sequence of networks were needed, and each of them was trained independently. Zhang \etal~\cite{zhang2018adversarial} later employed AE for weakly supervised object localisation, and trained two complementary classifiers jointly with localisation maps being online inferred. AE is adopted in this work to enhance food recognition. To this end, a completely different model from the ones used by the above two approaches is developed. In addition, we calculate CAM in an online manner to enable the end-to-end training. There are several other works sharing the similar idea with AE. For example, Singh and Lee~\cite{singh2017hide} proposed to hide patches randomly in an input image to boost weakly supervised object localisation. For the same purpose, Kim \etal~\cite{kim2017two} adopted a two-phase learning strategy of suppressing the second network's activation conditioned on the outcome of the first network.

\section{Method}\label{sec: method}

We use AE to progressively mine discriminative regions from an input food image. The representations of the mined regions and that of the original input image are concatenated as the final representation for food recognition. Our network architecture denoted as PAR-Net has three sub-networks (\ie, a primary network, an auxiliary network and a region network) and an extra fully connected layer. We use CAM in an online fashion to highlight the discriminative region at each mining step, which is described in Section~\ref{subsec: cam}. The network architecture as well as the training and inference procedures are detailed in Section~\ref{subsec: par_net}.

\subsection{Calculating CAM}\label{subsec: cam}
To calculate CAM~\cite{zhou2016learning}, a classification network normally follows the structure of {\tt conv}-{\tt gap}-{\tt fc}-{\tt softmax} on top. The generic way of producing CAM thus is defined as in Eqn.~\ref{eq: cam}. 

\begin{equation}\label{eq: cam}
    CAM(I, c) = \sum_{k = 1}^{N} w_{k,c} \cdot F_{k} 
\end{equation}

\noindent
where $c$ denotes the target class of an image $I$; $F_{k}$ is the $k^{th}$ feature map ($N$ in total) of the last convolutional layer; $w_{k,c}$ is the weight in the fully connected layer that represents the contribution of the $k^{th}$ neuron of the global average pooling (GAP) layer makes for identifying the class $c$ for $I$. 

We integrate CAM into both network training and inference, and calculate it differently during these two procedures. Specifically, in {\bf training}, $c$ is always adopted as the ground truth label associated with the image $I$.  Whereas in {\bf inference}, $c$ is the predicted top-1 class.

\subsection{PAR-Net}\label{subsec: par_net}

The network structure consists of four major parts: (\romannumeral 1) a primary network (P-Net) which classifies the original full input image; (\romannumeral 2) an auxiliary network (A-Net) which classifies images with discriminative region(s) erased; (\romannumeral 3) a region network (R-Net) which classifies the cropped and upsampled discriminative regions; (\romannumeral 4) an extra fully connected layer which classifies the concatenated representations of the input full image and the mined regions. An overview of the network architecture is shown in Figure~\ref{fig: net_structure}. The training and inference procedures are designed as follows:

\begin{figure}
  \includegraphics[width=\linewidth]{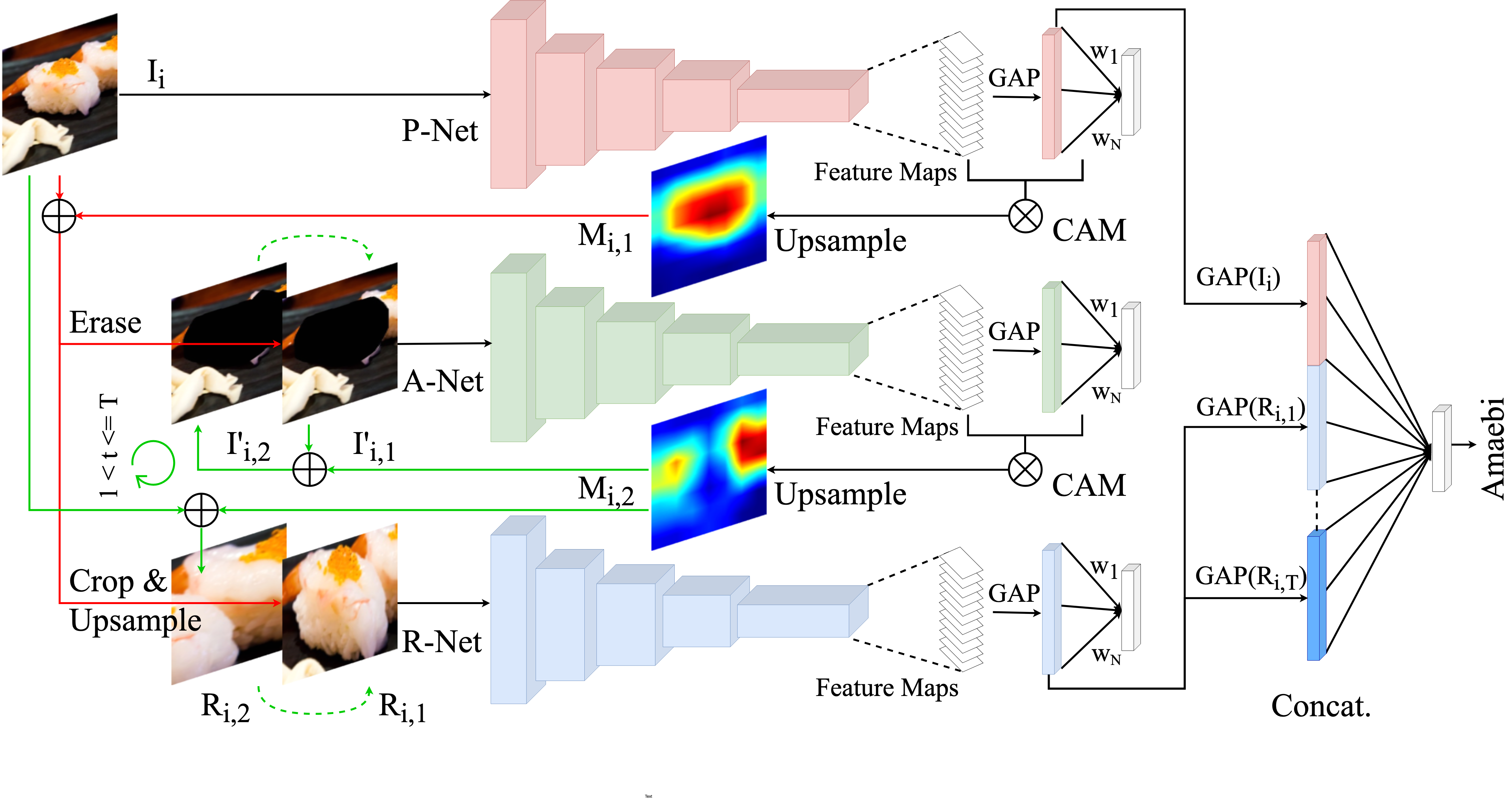}
  \caption{Overview of the network structure. The proposed PAR-Net consists of three sub-networks and an extra fully connected layer. The final prediction of an input food image $I_{i}$ is based on the concatenated representations of the input full image and the mined discriminative regions. Note that the last discriminative-region-erased image $I_{i, T}'$ does not need to be calculated as its classification loss is never accumulated during training.}
  \label{fig: net_structure}
\end{figure}

\noindent
{\bf Training.} We adopt the standard cross-entropy loss for all the classification involved during training. Concretely, given an image-label pair $\{I_{i}, y_{i}\}$ as the input, and the number of times of region mining $T$, we first feed $I_{i}$ into the P-Net for classification and denote the resulting loss as $\mathcal{L}_{p}$. In the meantime, the corresponding class activation map $CAM(I_{i}, y_{i})$ is calculated based on Eqn.~\ref{eq: cam}, and upsampled to obtain a heatmap indicating the discriminative region. In the following, we use $M_{i, t}$, $R_{i, t}$ and $I_{i, t}'$ to denote the CAM-based heatmap,  the mined region, and the image with the discriminative region(s) erased at mining step $t$ ($t \leq T$). All the $R_{i, t}$ and $I_{i, t}'$ inherit the ground truth label $y_{i}$ from the input image $I_{i}$.

Therefore, after upsampling $CAM(I_{i}, y_{i})$ to the same size as $I_{i}$, the first heatmap $M_{i, 1}$ is obtained. We threshold $M_{i, 1}$ to keep the values that are above $\alpha \times max(M_{i, 1})$, and then find the connected components. For each connected component, we sum all values inside. The one with the largest sum is used to indicate the most discriminative region in $I_{i}$. This operation for finding the discriminative region is slightly different from the original CAM implementation for object localisation ~\cite{zhou2016learning} since we are more interested in the discriminativeness than the localisation accuracy. We denote the most discriminative region highlighted by $M_{i, 1}$ in $I_{i}$ as $R_{i, 1}$, which could be of any shape. A tight bounding box covering $R_{i, 1}$ is then used to crop it out. We upsample the cropped patch using bilinear interpolation to the same size as $I_{i}$, and feed it into the R-Net for recognition. Simultaneously, we replace the pixels inside $R_{i, 1}$ in $I_{i}$ with zeros (as the first erasing operation),  resulting in the first discriminative-region-erased image $I_{i, 1}'$ which is then fed into the A-Net for classification. Note that the P-Net is only responsible for classifying the original full image and finding the first discriminative region. The remaining regions are mined by the A-Net. This is mainly to ensure that the accuracy of classifying a full input image can be preserved such that the input image's extracted representation is of high discriminativeness.

If $T > 1$, the A-Net then continues to discover the next discriminative region by calculating $CAM(I_{i, t}', y_{i}), t \in \{1, ..., T - 1\}$ while classifying $I_{i, t}'$ (the A-Net classifies $I_{i, t}'$ in an adversarial manner as $t$ increases, the $I_{i, t}'$ will have less discriminative regions left for the A-Net to rely on for identifying the correct class), and the R-Net keeps recognising the new discriminative region $R_{i, t}, t \in \{2, ..., T\}$ fed from the A-Net. It is worth noting that $R_{i, t+1}$ is always cropped from the original input image $I_{i}$ based on $M_{i, t+1}$ and the same ratio $\alpha$ for calculating the threshold $\alpha \times max(M_{i, t+1})$, and $I_{i, t+1}'$ is always obtained by erasing $R_{i, t+1}$ from $I_{i, t}'$. For each classification of the A-Net and the R-Net at each mining step $t$, we denote their losses as $\mathcal{L}_{a, t}$ and $\mathcal{L}_{r, t}$, respectively. We extract the representation of the input image $I_{i}$ and each region $R_{i, t}, t \in \{1, ..., T\}$ from the GAP layer of the P-Net and the R-Net, respectively. These representations are concatenated and input into the additional fully connected layer for classification. We use $y_{i}$ to calculate the loss, and denote it as $\mathcal{L}_{concat}$. Thus, depending on the number of times of region mining $T$, the total loss of the whole model is defined as in Eqn.~\ref{eq: loss}. 

\begin{equation}\label{eq: loss}
    \mathcal{L} = 
    \begin{cases}
        \mathcal{L}_{p}, & \text{if } T = 0 \\
        \mathcal{L}_{p} + \mathcal{L}_{r, 1} + \mathcal{L}_{concat}, & \text{if } T = 1\\
        \mathcal{L}_{p} + \sum_{t=1}^{T} \mathcal{L}_{r, t} + \mathcal{L}_{concat} + \sum_{t=1}^{T-1} \mathcal{L}_{a, t}, & \text{if } T > 1
    \end{cases}
\end{equation}

Notice that when $T = 0$, the overall loss $\mathcal{L}$ is equivalent to the loss $\mathcal{L}_{p}$ (\ie, classify the input image $I_{i}$ only by the P-Net), and when $T = 1$, the A-Net's loss $\mathcal{L}_{a, 1}$ is not counted. As a matter of fact, loss $\mathcal{L}_{a,t}$ is accumulated only if a discriminative region is mined from $I_{i,t}'$. The PAR-Net as a whole is then trained end-to-end.

\noindent
{\bf Inference.} In inference, since only an image $I_{i}$ is provided as the input, the calculation of CAM is different from that in training as mentioned in Section~\ref{subsec: cam}. The other parts however remain the same. To be more specific, when $T \geq 1$, the P-Net still mines the first discriminative region but instead uses $CAM(I_{i}, c_{i})$ where $c_{i}$ is the predicted top-1 class of the input image $I_{i}$ by the P-Net, and the A-Net mines the rest discriminative regions with $CAM(I_{i, t}', c_{i, t}), t \in \{1, ..., T - 1\}$ where $c_{i, t}$ is the estimated top-1 class of the image $I_{i, t}'$ by the A-Net. All mined regions are fed into the R-Net to obtain their representations. The output of the additional fully connected layer is used as the final prediction $\hat y_{i}$ for the input food image $I_{i}$.

\section{Experiments}

In this section, we first introduce the datasets used to validate the proposed approach, and then describe the implementation details. The results are then presented followed by the comparison between our approach and other state-of-the-arts. The ablation studies are reported at the end.

\subsection{Datasets}

The proposed method was evaluated on the following three food datasets:

\noindent
{\bf Food-101. } Food-101~\cite{bossard2014food} contains 101 food classes and each class has 1,000 images. The whole dataset is split into 75,750 images for training and 25,250 images for testing.  

\noindent
{\bf Vireo-172. } Vireo-172~\cite{chen2016deep} is a large scale Chinese food dataset with 172 food classes and 110,241 food images in total. The dataset is divided into training, validation, and test sets. Each class assigns $60\%$ of its images for training, $10\%$ for validation, and the rest $30\%$ for testing. 

\begin{figure}[!tbp]
  \centering
  \begin{minipage}[b]{0.48\textwidth}
    \includegraphics[width=\textwidth]{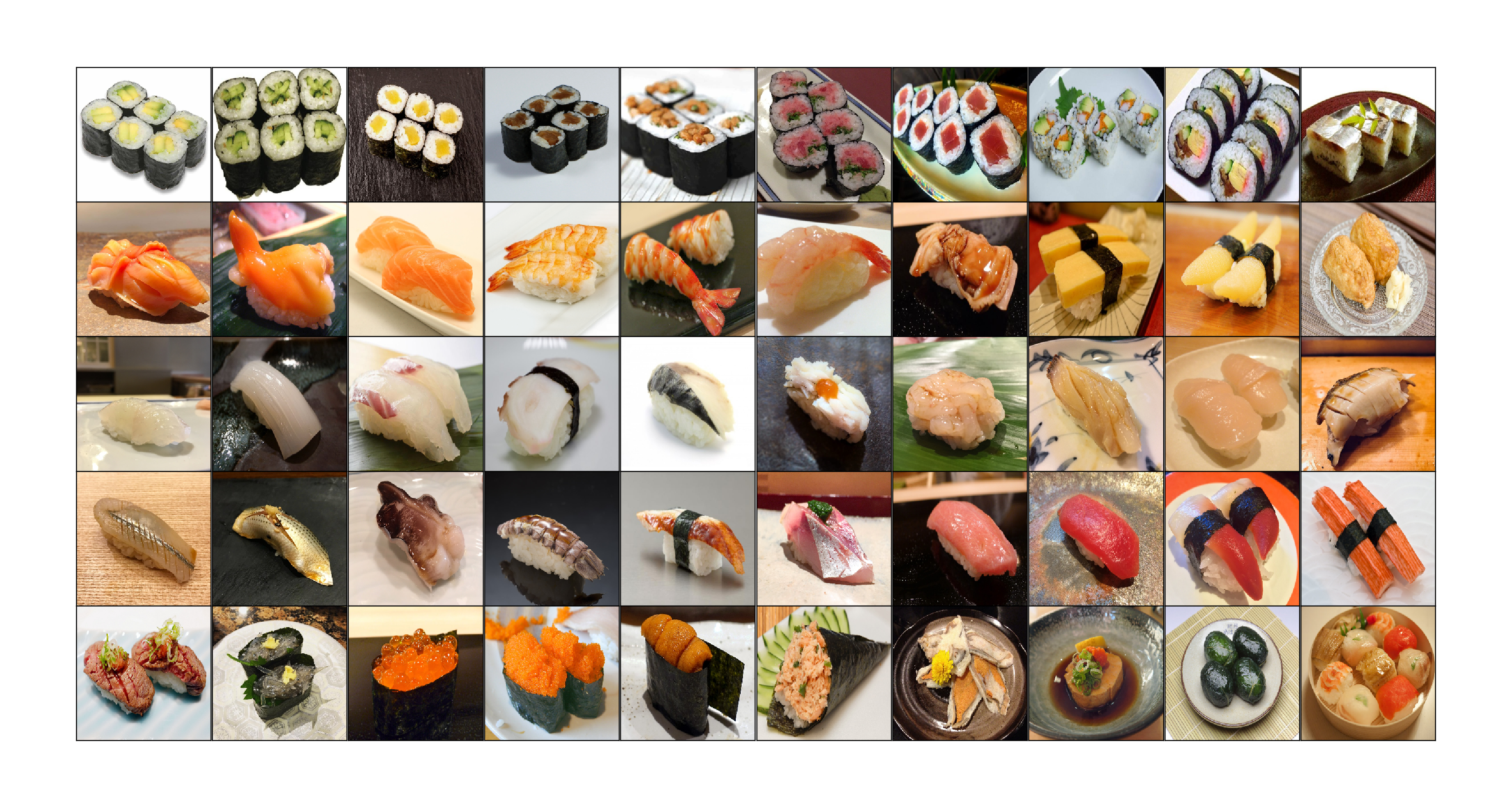}
    \caption{Samples of the Sushi-50 dataset (one sample is shown for each class)}
    \label{fig: sushi_samples}
  \end{minipage}
  \hfill
  \begin{minipage}[b]{0.5\textwidth}
    \includegraphics[width=\textwidth]{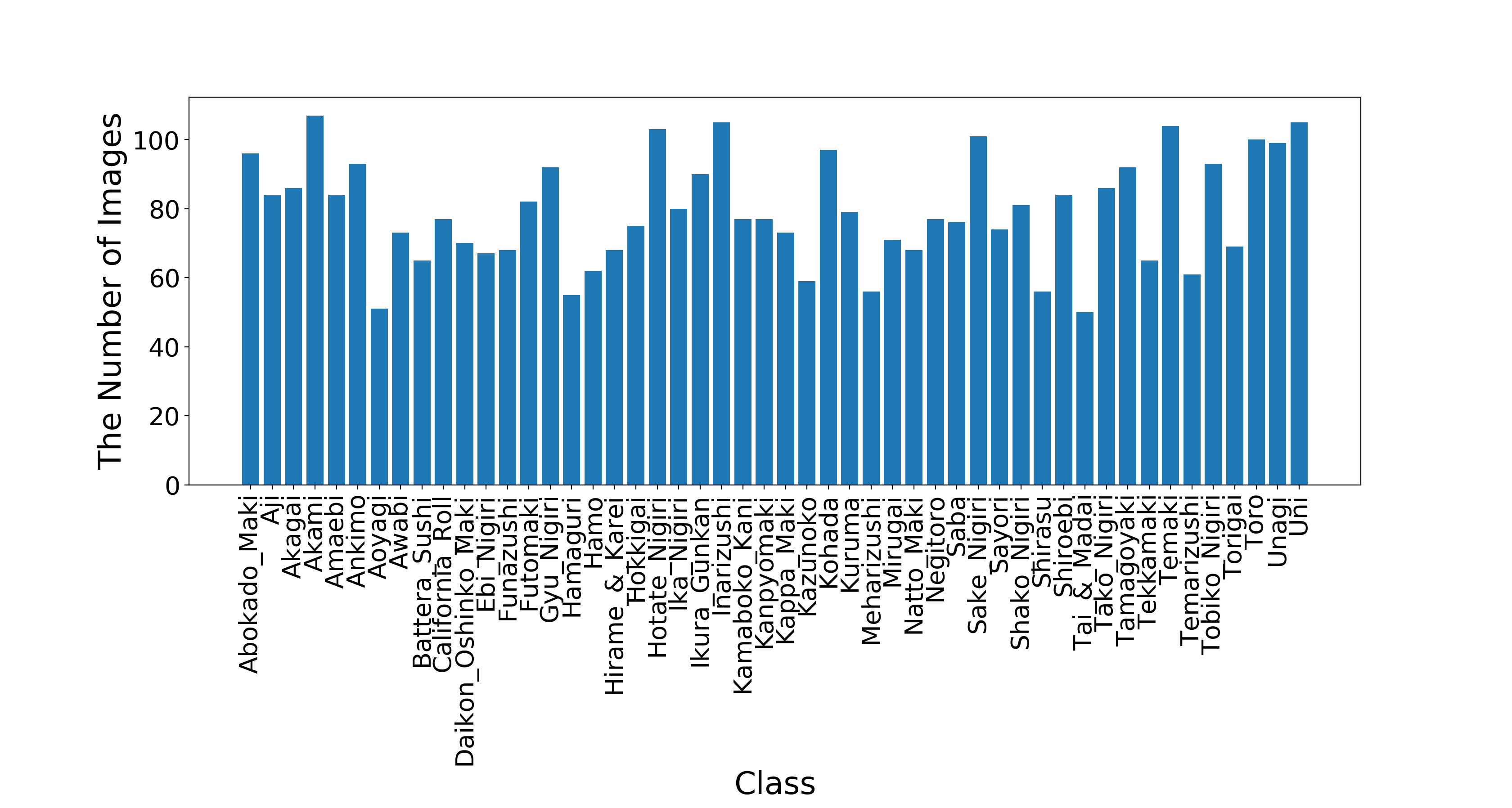}
    \caption{The number of images of each category of the Sushi-50 dataset}
    \label{fig: sushi_statistics}
  \end{minipage}
\end{figure}

\noindent
{\bf Sushi-50. } We built Sushi-50 \footnote{\url{http://www.doc.ic.ac.uk/~jq916/Sushi-50.zip}} as a fine-grained food dataset, which has 50 different sushi classes. These classes are selected based on a sushi guide \footnote{\url{https://www.japan-talk.com/jt/new/sushi-list}} with those not having sufficient images excluded. All images were downloaded from google and duplicates were removed. This new dataset has 3,963 images. Each class has $50\%$ of its images assigned for training and another $50\%$ retained for testing. Figure~\ref{fig: sushi_samples} shows one sushi sample for each class and Figure~\ref{fig: sushi_statistics} shows the number of images of each class.

\subsection{Implementation Details}\label{subsec: implementation}

Each sub-network within the PAR-Net is based on the ResNet~\cite{he2016deep}. We attempt various combinations of the sub-networks by using ResNets with different depth. In training, a crop is randomly sampled from the original image and resized to $224 \times 224$ with scale and aspect ratio augmentation~\cite{szegedy2015going} as the input, or its horizontal flip. No any further data augmentation technique is used. In inference, we use the center crop (1-crop) of size $224 \times 224$ from the original image (resized to $256 \times 256$). Following the practice in~\cite{martinel2018wide}, we also test our approach with the standard ten crops (10-crop). The PAR-Net is implemented with PyTorch. We initialise each sub-network with the weights of the corresponding ResNet model pretrained on ImageNet~\cite{deng2009imagenet} and fine-tune all layers afterwards. We train the PAR-Net for up to 35 epochs using stochastic gradient descent (SGD) with a momentum of 0.9, and a mini-batch size of 16. The initial learning rate is set to $10^{-3}$ and divided by 10 every 7 epochs. A weight decay of $10^{-4}$ is adopted. We set the ratio $\alpha$ to 0.5 and the number of times of region mining $T$ to 3. More detailed analysis about the choice of $\alpha$ and $T$ is given in Section~\ref{subsec: ablation}.

\subsection{Results}

\begin{table}[t]
\centering
\resizebox{\textwidth}{!}{%
\begin{tabular}{@{}lcccccc@{}}
\toprule
\multirow{2}{*}{Method}    & \multicolumn{2}{c}{Food-101 (Top-1)} & \multicolumn{2}{c}{Vireo-172 (Top-1)} & \multicolumn{2}{c}{Sushi-50 (Top-1)} \\ \cmidrule(l){2-7} 
                           & 1-crop            & 10-crop          & 1-crop            & 10-crop           & 1-crop            & 10-crop          \\ \midrule
ResNet-50                  & 87.1              & 88.2             & 87.0              & 87.7              & 88.9              & 89.0             \\
ResNet-101                 & 88.1              & 89.0             & 87.5              & 88.3              & 89.7              & 90.0             \\
PAR-Net (P50+A34+R50)      & 88.5              & 89.5             & 88.8              & 89.3              & 91.2              & 91.0             \\
PAR-Net (P101+A34+R50)     & \textbf{89.3}     & 90.2             & 89.2              & 89.7              & \textbf{92.3}     & 91.9             \\
PAR-Net (P101+A101+R101)* & \textbf{89.3}     & \textbf{90.4}    & \textbf{89.6}     & \textbf{90.2}     & 91.8              & \textbf{92.0}    \\ \bottomrule
\end{tabular}%
}
\caption{Top-1 testing accuracy (\%) on the three food datasets. ResNet-50 and ResNet-101 are based on our test. The PAR-Net is tested with different combinations of its subnets (\eg, P50 denotes the P-Net is based on ResNet-50). The best accuracy of both 1-crop and 10-crop is written in bold. All the PAR-Net versions reported mine 3 discriminative regions and use a ratio $\alpha$ of 0.5. * was trained with a batch size of 8 for up to 60 epochs, and the learning rate was divided by 10 every 14 epochs.}
\label{tab: overall_results}
\end{table}

The overall results on the test sets are shown in Table~\ref{tab: overall_results}. The accuracy of the PAR-Net is the accuracy of classifying the concatenated representation. The PAR-Net with its P-Net and R-Net being two individual ResNet-50s and its A-Net being a ResNet-34 (represented as P50+A34+R50 in Table~\ref{tab: overall_results}) improves the 1-crop testing accuracy by 1.4\%, 1.8\%, 2.3\% on Food-101, Vireo-172, and Sushi-50, respectively, compared to the baseline results of using a single ResNet-50. Similar improvements can also be observed in the 10-crop testing. When the P-Net is instantiated with a ResNet-101, the accuracy on all the three datasets further increases. We achieve the best 10-crop accuracy on all the datasets when using ResNet-101 for each sub-network.

\begin{table}[t]
\parbox[t]{.5\linewidth}{
\centering
\begin{tabular}[t]{@{}lc@{}}
\toprule
Method                   & Top-1         \\ \midrule
RFDC~\cite{bossard2014food}          & 50.76         \\
DCNN-FOOD~\cite{yanai2015food}    & 70.41         \\
DeepFood~\cite{liu2016deepfood}     & 77.4          \\
Inception V3~\cite{hassannejad2016food} & 88.28         \\
DLA (CVPR2018)~\cite{yu2018deep}           & 90.0          \\
WISeR (WACV2018)~\cite{martinel2018wide}         & 90.27         \\
DSTL (CVPR2018)~\cite{cui2018large}          & \textbf{90.4} \\ \midrule
PAR-Net (Ours)           & \textbf{90.4} \\ \bottomrule
\end{tabular}
\caption{Results compared with other state-of-the-art methods on Food-101}
\label{tab: food101_camparison}
}
\hfill
\parbox[t]{.5\linewidth}{
\centering
\begin{tabular}[t]{@{}lc@{}}
\toprule
Method             & Top-1         \\ \midrule
VGG~\cite{Simonyan15}    & 80.41         \\
Arch-D (ACMMM2016)~\cite{chen2016deep} & 82.06         \\ \midrule
PAR-Net (Ours)     & \textbf{90.2} \\ \bottomrule
\end{tabular}
\caption{Comparison on Vireo-172}
\label{tab: vireo172_comparison}
\centering
\begin{tabular}{@{}lc@{}}
\toprule
Method                & Top-1         \\ \midrule
ResNet-101~\cite{he2016deep} & 90.0          \\ \midrule
PAR-Net (Ours)        & \textbf{92.0} \\ \bottomrule
\end{tabular}
\caption{Comparison on Sushi-50}
\label{tab: sushi50_comparison}
}
\end{table}

\begin{figure}[t]
  \centering
  \includegraphics[width=0.9\linewidth]{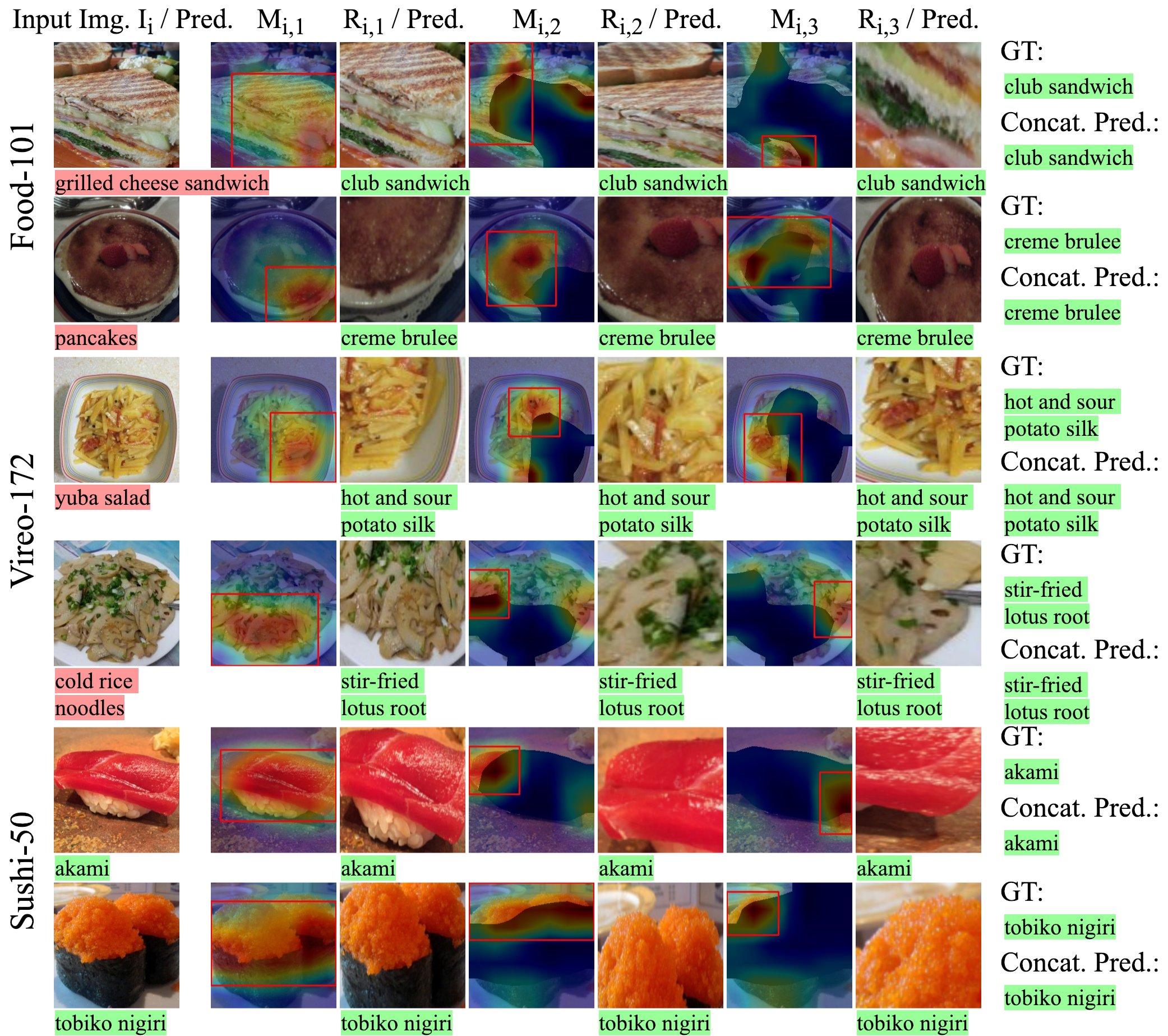}
  \caption{visualisation of the predicted results of the PAR-Net (P101+A101+R101) on the test sets. Two examples are shown for each food dataset. The CAM-based heatmap $M_{i, t}$ has been projected onto the image it is calculated from. Each discriminative region $R_{i, t}$ has been cropped using the bounding box (drawn as red in $M_{i, t}$) and upsampled to the same size as the input $I_{i}$ (refer to Section~\ref{subsec: par_net}). The shaded area in $M_{i,2}$ and $M_{i,3}$ are the (accumulated) erased region(s). We use pink to indicate the incorrectly predicted class and green the correctly predicted as well as the ground truth class. Input Img. $I_{i}$ is classified by the P-Net and $R_{i, t}$ the R-Net. Concat. Pred. is the prediction based on the concatenated representation.}
  \label{fig: vis_samples}
\end{figure}

We then compare our approach with other state-of-the-arts. The result achieved by our approach on Food-101 (see Table~\ref{tab: food101_camparison}) outperforms~\cite{martinel2018wide} in which more data augmentation techniques are used, and matches~\cite{cui2018large} whereas in~\cite{cui2018large} higher image resolution and a more sophisticated transfer learning approach are adopted. On Vireo-172, our approach achieves the current best accuracy (see Table~\ref{tab: vireo172_comparison}), and on Sushi-50, compared to a baseline method, the proposed approach yields better performance (see Table~\ref{tab: sushi50_comparison}). We visualise some predicted results of the PAR-Net on the test sets (two examples for each dataset) in Figure~\ref{fig: vis_samples}. In the first four rows, the input food image is incorrectly recognised by the P-Net. However, the following mined discriminative regions are successfully classified by the R-Net. The final predication based on the concatenated representation therefore is correct. In the bottom two rows, the input as well as the mined regions are correctly recognised by the P-Net and the R-Net, respectively, which leads to the desired prediction after concatenating representations. Diversity of the mined discriminative regions within each image can also be observed in Figure~\ref{fig: vis_samples}, which is beneficial for the more accurate final prediction. The results obtained demonstrate the effectiveness of using mined discriminative regions to improve food recognition and show consistent performance of our method across multiple food datasets.

\subsection{Ablation Studies}\label{subsec: ablation}

\begin{table}[t]
\centering
\resizebox{\textwidth}{!}{%
\begin{tabular}{@{}cccccccccc@{}}
\toprule
\multirow{2}{*}{Dataset} & \multicolumn{3}{c}{T}                                     & \multicolumn{3}{c}{$\alpha$}                            & \multicolumn{3}{c}{Network Structure}                           \\ \cmidrule(l){2-10} 
                         & 2    & 3             & 4                                  & 0.3  & 0.5           & 0.7                       & No weights sharing & P\&R share weights & P\&A\&R share weights \\ \midrule
Vireo-172                & 88.5 & 88.8          & \multicolumn{1}{c|}{\textbf{89.0}} & 88.7 & \textbf{88.8} & \multicolumn{1}{c|}{87.8} & \textbf{88.8}      & 88.1               & 87.3                  \\
Sushi-50                 & 90.5 & \textbf{91.2} & \multicolumn{1}{c|}{91.0}          & 90.3 & \textbf{91.2} & \multicolumn{1}{c|}{91.1} & \textbf{91.2}      & 90.6               & 89.8                  \\ \bottomrule
\end{tabular}%
}
\caption{Ablation analysis of the impact of the number of times of region mining $T$, the ratio $\alpha$, and the different network structure on the accuracy (\%) of food recognition.}
\label{tab: abl-1}
\end{table}

\begin{table}[t]
\centering
\resizebox{\textwidth}{!}{%
\begin{tabular}{@{}lcccccccccccccc@{}}
\toprule
\multirow{2}{*}{Dataset}          & \multicolumn{7}{c}{PAR-Net (P50+A34+R50)}                                    & \multicolumn{7}{c}{PAR-Net (P101+A101+R101)}            \\ \cmidrule(l){2-15} 
          & $I$    & $R_{1}$    & $R_{2}$    & $R_{3}$    & $I_{1}'$    & $I_{2}'$    & $concat$                                  & $I$    & $R_{1}$    & $R_{2}$    & $R_{3}$    & $I_{1}'$    & $I_{2}'$    & $concat$             \\ \midrule
Food-101  & 87.1 & 86.8 & 84.0 & 80.4 & 65.6 & 50.7 & \multicolumn{1}{c|}{\textbf{88.5}} & 87.8 & 87.4 & 82.3 & 73.8 & 79.1 & 68.1 & \textbf{89.3} \\
Vireo-172 & 87.0 & 87.6 & 85.0 & 81.1 & 60.5 & 37.2 & \multicolumn{1}{c|}{\textbf{88.8}} & 88.0 & 87.6 & 82.7 & 74.9 & 77.1 & 60.4 & \textbf{89.6} \\
Sushi-50  & 88.9 & 90.6 & 87.1 & 81.7 & 42.2 & 20.6 & \multicolumn{1}{c|}{\textbf{91.2}} & 89.0 & 90.2 & 83.2 & 75.2 & 52.8 & 27.0 & \textbf{91.8} \\ \bottomrule
\end{tabular}%
}
\caption{Average top-1 testing accuracy (\%) on the full input image ($I$), discriminative region $R_{t}$ and region-erased image $I_{t}'$ of mining step $t$, and the concatenated representation. Two different PAR-Nets' performances are shown when mining with 3 discriminative regions.}
\label{tab: abl-2}
\end{table}

{\bf Impact of the number of times of region mining. } To study how many regions should be mined so as to achieve the optimal result, we conducted experiments with the PAR-Net (P50+A34+R50) on Vireo-172 and Sushi-50. As shown in the left of Table~\ref{tab: abl-1}, mining 3 discriminative regions in general is sufficient for achieving satisfactory performance. The accuracy drops when less regions are mined. Although the accuracy is slightly higher on Vireo-172 when the number of mined regions increases to 4, it inevitably requires more overhead in training and inference. Therefore, we adopt $T = 3$ for most of our experiments.

\noindent
{\bf Impact of the ratio used to calculate the threshold. } We threshold a CAM-based heatmap by using a ratio $\alpha$ multiplied with the maximum value of the heatmap, after which we use the connected component with the largest sum to indicate the most discriminative region. The $\alpha$ therefore determines the size of the discriminative region, the smaller the $\alpha$, the larger the size. We show the impact of the value of $\alpha$ on the accuracy of food recognition in the middle of Table~\ref{tab: abl-1}. The experiments are based on using the PAR-Net (P50+A34+R50) and a fixed $T = 3$. In general, $\alpha = 0.5$ works well on both Vireo-172 and Sushi-50. However, the results suggest that the choice of $\alpha$ could be dataset-dependent as $\alpha = 0.3$ is only marginally worse than $\alpha = 0.5$ whereas $\alpha = 0.7$ causes a clear drop in the accuracy on Vireo-172. An opposite trend can be observed on Sushi-50. 

\noindent
{\bf Impact of the network structure. } So far, the accuracy reported of the PAR-Net is based on using three individual sub-networks. As both P-Net and R-Net learn food representations (global and local, respectively), an intuition is that they could share weights. Therefore, we used a single network (ResNet-50) to replace the P-Net and R-Net, and still kept the A-Net (ResNet-34). This modified structure was trained with the same setup as in Section~\ref{subsec: implementation}. It can be observed from the right of Table~\ref{tab: abl-1} that this structure is of inferior performance compared to the original PAR-Net (P50+A34+R50). When we forced all three sub-networks to share weights, \ie, replacing them with a single ResNet-50, the accuracy further decreases. This can be interpreted as a limitation of the PAR-Net that it is necessary to have independent sub-networks to conduct different tasks in order to achieve good performance.

\noindent
{\bf Accuracy of each sub-network. }As classification occurs in each sub-network (the P-Net classifies the input image, the R-Net classifies the mined regions, and the A-Net classifies discriminative-region-erased images), we show each sub-network's classification accuracy on the corresponding target, which is summarised in Table~\ref{tab: abl-2}. As mining continues, the accuracy of classifying the mined region decreases as expected ($R_{1} > R_{2} > R_{3}$), because the later mined region is less discriminative than the earlier ones. This is also true for the images with discriminative region(s) erased, the accuracy degrades ($I_{1}' > I_{2}'$) as the later image have less discriminative regions left compared to the previous one. It is worth noting that the accuracy based on the concatenated representation is always higher than that of the input image and the mined regions, which verifies that the integration of global and local representations contributes to better food recognition.

\section{Conclusions}

We introduced a novel model for food recognition, which progressively mines discriminative food regions and merges their representations with the full input image's to make an accurate prediction. The model has been validated on multiple food datasets including a new fine-grained one introduced by this paper, and yields state-of-the-art performance. By providing more accurate recognition results, the proposed model is expected to facilitate the development of passive dietary monitoring. Investigating vision-based approaches for food ingredient level analysis is planned in future work.

\noindent
{\bf Acknowledgements. }This work is supported by the Innovative Passive Dietary Monitoring Project funded by the Bill \& Melinda Gates Foundation (Opportunity ID: OPP1171395).

\end{document}